\documentclass{article}

\usepackage{algorithm}
\usepackage{algorithmic}
\usepackage{color}
\usepackage{svg}
\usepackage{expl3}
\usepackage{adjustbox}
\usepackage{bm}
\usepackage{multirow}

\usepackage{changes}

\usepackage[final]{neuripsBDL_2021}

\usepackage[utf8]{inputenc} 
\usepackage[T1]{fontenc}    
\usepackage{hyperref}       
\usepackage{url}            
\usepackage{booktabs}       
\usepackage{amsfonts}       
\usepackage{amsmath, amssymb, amsthm}
\usepackage{xfrac}
\usepackage{nicefrac}       
\usepackage{microtype}      
\usepackage{xcolor}         

\usepackage{array,multirow}
\usepackage{lipsum}

\makeatletter

\title{An Empirical Comparison of GANs and\\ Normalizing Flows for Density Estimation}

\author{%
    Tianci Liu\\
      Purdue University\\
      \texttt{liu3351@purdue.edu}
      \And
      Jeffrey Regier\\
      University of Michigan\\
      \texttt{regier@umich.edu}
}

\begin{document}
\maketitle









\begin{abstract}
Generative adversarial networks (GANs) and normalizing flows are both approaches to density estimation that use deep neural networks to transform samples from an uninformative prior distribution to an approximation of the data distribution. There is great interest in both for general-purpose statistical modeling, but the two approaches have seldom been compared to each other for modeling non-image data. The difficulty of computing likelihoods with GANs, which are implicit models, makes conducting such a comparison challenging. We work around this difficulty by considering several low-dimensional synthetic datasets. An extensive grid search over GAN architectures, hyperparameters, and training procedures suggests that no GAN is capable of modeling our simple low-dimensional data well, a task we view as a prerequisite for an approach to be considered suitable for general-purpose statistical modeling. Several normalizing flows, on the other hand, excelled at these tasks, even substantially outperforming WGAN in terms of Wasserstein distance---the metric that WGAN alone targets. Scientists and other practitioners should be wary of relying on WGAN for applications that require accurate density estimation.
\end{abstract}

\section{Introduction}

Generative adversarial networks (GANs) and normalizing flows can be seen as alternative approaches: both are flexible generative models that, in contrast to both variational autoencoders and traditional ``shallow'' Bayesian models, do not assume that either the likelihood or the prior has a simple parametric form.
This flexibility makes both approaches appealing for modeling complex scientific data. GANs in particular have caught the attention of scientists~\citep{mustafa2019cosmogan, Choi2017GeneratingMD,10.1093/jamia/ocy142}.

To date, however, GANs have been validated primarily using image data.
Little research exists investigating whether GANs are suitable for general statistical modeling, as would be required for scientific applications.
Performance metrics used to assess GANs thus far have unfortunately only revealed half the story: they measure whether the data GANs generate are realistic (i.e., precision) but not whether the fitted GAN model has support for held-out samples (i.e., recall). The difficulty associated with measuring recall stems from the intractability of the likelihood GANs assign to high-dimensional data---a well known limitation of implicit models~\cite{lucic2018are}.

Synthetic low-dimensional data, on the other hand, offers us the potential to establish a negative result. Using these data, we can accurately assess the performance of both GANs and flows. We study the performance of both GANs and flows on synthetic univariate data from  mixture data (Section~\ref{data}). Although accurately learning distributions of univariate data is not sufficient for scientific modeling, it is a necessary condition for a tool to be reliable.

Even with univariate data, measuring performance is not trivial. We confront subtle issues of selecting kernel density estimation bandwidth, and present a low-variance estimator for Wasserstein distance between low dimension distributions (Section~\ref{metrics}). The visualization methods we develop allow us to demonstrate several distinct failure modes of GANs.

An additional challenge of establishing negative results with respect to GANs in general is their diversity: there are hundreds of different GAN algorithms, each of which can be combined with numerous tricks and tweaks, implying an exponential number of combinations. Using eight NVIDIA GeForce RTX 2080 Ti GPUs, and weeks of runtime, we systematically search these combinations. For WGAN, we experiment with gradient penalties, spectral normalization, batch normalization, cyclic learning rates, ResNet architectures, various layer widths, different noise distributions, and additional tuning parameters (Section~\ref{methods}). We also experiment extensively with normalizing flows.

GANs failed to learn even basic structures in our synthetic data, whereas some normalizing flows modeled the data well~(Section~\ref{results}).
Surprisingly, normalizing flows outperform WGANs even in terms of the metric that only WGAN targets: minimizing Wasserstein-1 distance.

The implications of these experiments, conducted solely with low-dimensional data, may not generalize to higher dimensions; hence, good performance on these experiments should be viewed necessary---but not sufficient---for a method to be suitable for general-purpose density estimation.
Scientists and other practitioners should be wary of relying on WGAN for applications that require accurate density estimation (Section~\ref{discussion}).


\section{Synthetic data}
\label{data}

We developed two synthetic datasets for evaluating GANs and normalizing flows.
Both datasets are univariate. To avoid confounding our results with issues of data efficiency and overfitting, which are beyond the scope of this work, we make the size of both datasets effectively infinite by drawing fresh data at every step.

\subsection{Unimodal dataset}
\label{data: gum}
One data model we consider is a two-component univariate mixture with equal means, specified by the following generative process:
\begin{align}
    y &\sim \mathrm{Bernoulli}(p) \\
    x \mid y=0 &\sim \mathrm{Uniform}(\mu - r\sigma, \mu + r\sigma) \\
    x \mid y=1 &\sim \mathrm{Gaussian}(\mu, \sigma^2).
\end{align}
Here $y$ is an unobserved random variable and $x$ is the data.
We set $p = 0.75, \mu=5, \sigma=0.1, r=5$. 
Figure~\ref{fig:experiment} (a) shows this density.
The density has several qualitative aspects we expect a good model to recover: sharply changing density at 4.5 and 5.5, and a bell-shaped mode symmetry around 5.

\subsection{Multimodal dataset}
\label{data: mgum}
We also consider a mixture of unimodal distributions with unequal means, as specified by the following generative process:
\begin{align}
    z &\sim \mathrm{Categorical}(\sfrac{1}{K}, \dots, \sfrac{1}{K}) \\
    y \mid z=k &\sim \mathrm{Bernoulli}(p) \\
    x \mid y=0 &\sim \mathrm{Uniform}(\mu_k - r\sigma, \mu_k + r\sigma) \\
    x \mid y=1 &\sim \mathrm{Gaussian}(\mu_k, \sigma^2).
\end{align}
Here $x$ is the data and $y$ and $z$ are unobserved random variables, used only to facilitate data generation.
We set $K = 8$, $p = 0.5$, $\sigma=2$, and $r=1.5$.
For $k=1,\ldots,K$, we set $\mu_k=10k$.
Figure~\ref{fig:experiment} (e) shows this density. We designed this density to have multiple modes, with non-negligible density between them. Compared to our unimodal mixture, our multimodal mixture has narrower modes with sharper boundaries.
The density has several qualitative aspects that we would expect a good model to recover: the correct number of modes, the correct high and low density areas, equal densities at the modes, and equal densities within the low-probability regions between modes.

\vspace{10px}
\section{Metrics}
\label{metrics}

Evaluating the quality of GANs is not trivial because the density of GAN generators cannot be evaluated directly---generators can only be sampled.
Our strategy is to first draw a large number of samples from the generator.
Because our data is low-dimensional, we can attain high sample density.
Then, we compute two metrics---one qualitative and the other quantitative--using these samples.

\subsection{Kernel density estimation}
Kernel density estimation (KDE) lets us visualize the GAN density and compare it qualitatively to the normalizing flow density and the true data generating density. However, KDE can be inaccurate if the bandwidths are chosen improperly: too large and the GAN appears smoother than it is, too small and the GAN density incorrectly appears to be highly variable. Either case can mask the extent to which a GAN captures structure in the true data distribution.

We strike a good balance by using a large sample size ($100,000$), and by choosing a bandwidth such that each band contains, on average, $500$ samples, where $500$ is chosen to approximately maximize the likelihood of held-out data.

\vspace{10px}
\subsection{Wasserstein-1 distance}
Wasserstein-1 distance, also known as Earth-Mover distance, measures the difference between two distributions. Wasserstein-1 distance is an especially interesting distance for our research questions because it is the objective function that WGAN targets.
While computing Wasserstein-1 distance in general requires solving a constraint optimization, for univariate data it can be readily estimated to high precision~\citep{ramdas2015wasserstein,panaretos2019statistical}.
Suppose $X$ (resp. $Y$) is a random variable following $\mathbb{P}_X$ (resp. $\mathbb{P}_Y$).
Let $x$ (resp. $y$) denote i.i.d. samples with size $n$ (resp. $m$) of $X$ (resp. $Y$) and let $\hat{F}_X $ (resp. $\hat{F}_Y$) be the empirical cumulative distribution function based on $x$ (resp. $y$). Then, Wasserstein-1 distance can be estimated with
\begin{align}
\label{directw1}
    W_1(\mathbb{P}_X, \mathbb{P}_Y) 
    &\approx \sum_{t \in x \cup y} |\hat{F}_X(t) - \hat{F}_Y(t)|.
\end{align}
This formula provides a convenient method of validating the predictions of a GAN critic $c$~\citep{arjovsky2017wasserstein, wgan-gp}, which estimates the Wasserstein-1 distance as
\begin{align}
\label{criticw1}
    W_1(\mathbb{P}_X, \mathbb{P}_Y) 
    &\approx \frac{1}{n}\sum_{i=1}^{n} c(x_i) - \frac{1}{m}\sum_{j=1}^{m} c(y_j).
\end{align}

\section{Methods}
\label{methods}

\begin{figure}
    \centering
    \includegraphics[width=\textwidth]{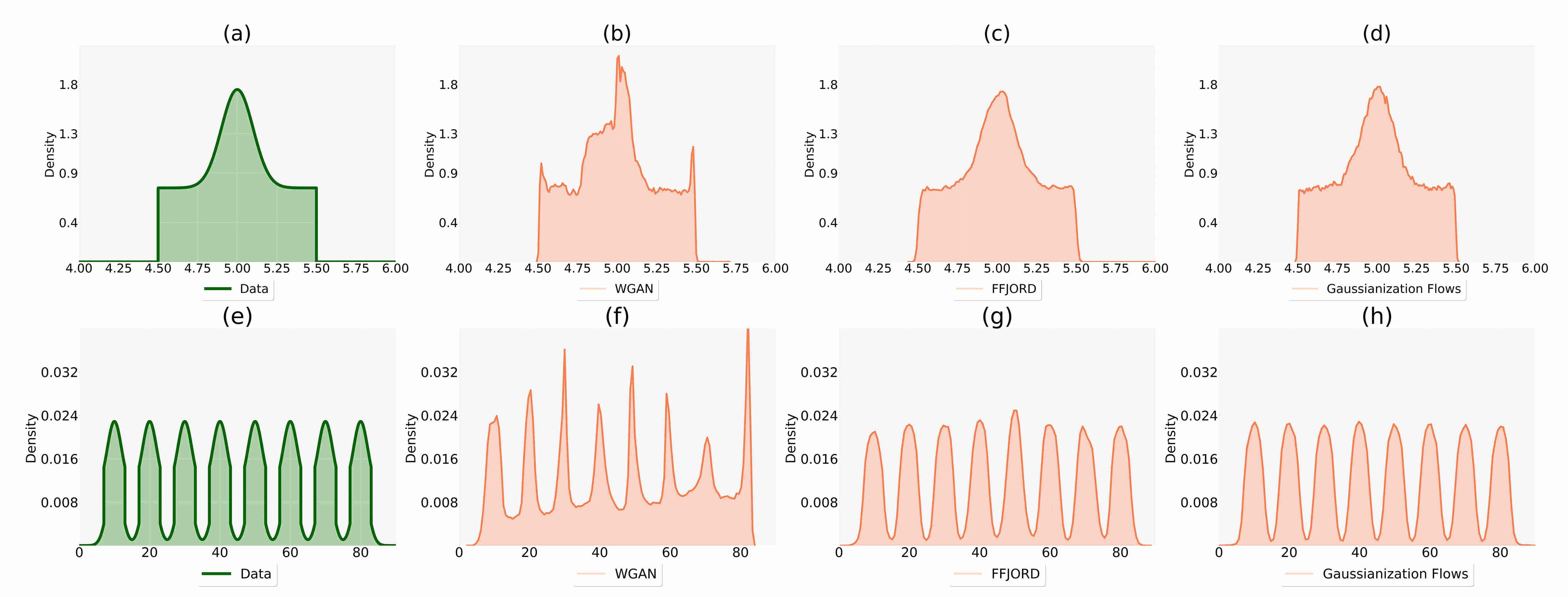}
    \caption{Density plots for our datasets and the models fitted to them. The top row pertains to the univariate dataset and the bottom row to the multivariate dataset. The leftmost column shows the data distributions. The other columns show the distributions learned by three density estimation algorithms: WGAN, FFJORD, and Gaussianization Flows.
    }
    \label{fig:experiment}
\end{figure}

We initially set out to demonstrate a positive result: that GANs are versatile statistical tools.
As evidence to the contrary accumulated, we embarked on the more challenging task of showing a negative result of some generality by searching large numbers of permutations of GAN types, architectures, and training techniques.
For normalizing flows, a positive result emerged fairly soon, so it was not necessary to try many combinations of flow training techniques and flow architectures.

\subsection{Methods for training GANs}
We considered the original GAN~\citep{goodfellow2014generative} at first, but focused on the Wasserstein GAN (WGAN) after it became apparent the former would not adequately model our data.
Prior work establishes that WGAN is among the best performing GANs if architectures are well-chosen~\citep{rosca2018distribution, lucic2018are,dinh2017density}.

The WGAN hyperparameters can be divided into two sets based on whether they determine network architecture or training strategy. 
The former includes activation functions, width, and depth, as well as different ways to initialize weights---namely, uniform and Xavier~\citep{pmlr-v9-glorot10a}. We also considered both fully connected and ResNet architectures~\citep{resnet}. For the generator, we considered both Gaussian and uniform priors of various dimensions, as well as batch normalization~\citep{batch_norm}. For the critic, we explored gradient penalties~\citep{wgan-gp} and spectral normalization~\citep{miyato2018spectral}, and consider different strengths for both regularizers.

The WGAN hyperparameters dictating the training strategy include optimizer learning rate, weight decay, and both coefficients for Adam~\citep{adam}. We also considered cyclic learning rates~\citep{cycliclr} following some failure modes we observed.
We also experimented with various numbers of critic updates per generator update. 

Exhaustive grid searching on such a large search space is prohibitively expensive. We approximate an exhaustive grid search with random search~\citep{10.5555/2188385.2188395} through ASHA~\citep{li2018massively}, which leverages early stopping scheduler of parallel hyperparameter tuning, to study the performances of different combinations.

\subsection{Methods for training flows}
We first considered Masked Autoregressive Flow~\citep{papamakarios2017masked}, Inverse Autoregressive Flow~\citep{NIPS2016_6581}, RealNVP~\cite{dinh2017density} and Planar flows~\cite{pmlr-v37-rezende15}. For univariate data such as ours, however, these flows have just two learnable scalar parameters. With so few learnable parameters, the capacity of such flows is limited for univariate data, so we did not consider them further.

Our experiments with flows focused instead on Free-form Jacobian of Reversible Dynamics (FFJORD)~\citep{grathwohl2018scalable}
and Gaussianization Flows (GF)~\citep{Meng2020GaussianizationF}.
The former, FFJORD, is a continuous flow based on an ordinary differential equation.
The latter, GF, stacks two types of learnable transformations: a linear transformation to rotate data such that correlations between different dimensions is minimized, and a non-linear transformation to learn each marginal distributions separately by composing inverse Gaussian CDF with a mixture of logistic distributions. The first type of transformation was unnecessary for our univariate data.
We did not run a grid search to find optimal parameters for either FFJORD or GF.
Instead, we used the same architectures and training strategies from~\citet{grathwohl2018scalable} and~\citet{Meng2020GaussianizationF}.


\section{Results}
\label{results}

We evaluated WGAN, FFJORD and Gaussianization Flows on both our synthetic datasets.
The results reported for WGAN are always for models tuned with extensive hyperparameter optimization.
Typically spectral normalization led to the best results for WGAN.
We consider the experimental results both qualitatively and quantitatively.



\subsection{Qualitative}

Figure~\ref{fig:experiment} is a key result of ours that shows the dataset densities (ground truth) and the learned densities. Only the best performing WGAN is shown. Even the best WGAN failed to reconstruct key qualitative features of the datasets.
On the unimodal dataset, WGAN assigned too much mass in both regions around the boundaries of data distribution's support, as well as at the mode of the data distribution. Further, WGAN failed to capture the symmetry and bell-shaped structure. On the multimodal dataset, WGAN recovered all modes and gaps between modes. However, WGAN misrepresented the local structure of the individual modes (e.g., the symmetry), as well as the relative densities of the modes, which should have been equal.

Both types of normalizing flows, in contrast, recovered both the local and global structure of the data distributions.
GF appears to be slightly more accurate than FFJORD.

\subsection{Quantitative}

Table~\ref{table:experiment} summarizes the quantitative performances on two datasets in terms of the Wasserstein-1 distance, as estimated by Equation~\ref{directw1}. Surprisingly, both flows outperformed the best WGAN in terms of Wasserstein-1 distance---a metric that only WGAN targets.

The best performing WGAN for unimodal data used spectral normalization to constrain the Lipschitz constant of the critic,
whereas for multimodal data, a gradient penalty worked better than spectral normalization.
Many of the modifications of the WGAN that we thought might help did not.
We report results for several of these modifications in Table~\ref{table:wgan}.

We made some progress in understanding why WGAN does not perform better.
Because our data is low dimensional, Equation~\ref{directw1}, gives us low variance (and unbiased) estimates of the Wasserstein-1 distance.
The WGAN critic computes Wasserstein distance differently, using Equation~\ref{criticw1}.
When WGAN converged, the Wasserstein-1 distance estimated by the critic often severely underestimated the true Wasserstein-1 distance.
Surprisingly, in some cases, the critic estimate of the Wasserstein-1 distance was negative.
Without reliable estimates from the critic of the distance between the data distribution and the model/generator density at the current iterate,
there is no sound basis for updates to the generator.

\begin{table}[ht!]
\caption{\textbf{Wasserstein-1 distance} between fitted models and the targeted data distributions (i.e., our unimodal dataset and our multimodal dataset), as estimated by Equation~\ref{directw1}.
\textbf{Lower is better.}
}
\label{table:experiment}
\begin{center}
\begin{small}
\begin{sc}
\begin{tabular}{lcc}
\toprule
{} & Unimodal Data &  Multimodal Data \\
\midrule
Wasserstein GAN & 0.0087 & 0.4814 
\\ FFJORD & 0.0066 & 0.289
\\Gaussianization Flows & \textbf{0.0035} & \textbf{0.138}
\\
\bottomrule
\end{tabular}
\end{sc}
\end{small}
\end{center}
\end{table}

\begin{table}[ht!]
\caption{\textbf{Wasserstein-1 distance} between fitted WGANs and the targeted data distributions, as estimated by Equation~\ref{directw1}.
\textbf{Lower is better.}
``Baseline'' is our best model: WGAN with either spectral normalization or gradient penalty, optimally tuned.
``With uniform prior'' substitutes a uniform prior for a Gaussian prior.
``With cyclic LR'' introduces a cyclic learning rate.
``With dropout'' adds dropout regularization.
``With ResNet'' uses residual blocks in the generator and the critic.
}
\label{table:wgan}
\begin{center}
\begin{small}
\begin{sc}
\begin{tabular}{lcc}
\toprule
{} & Unimodal Data &  Multimodal Data \\
\midrule
\textit{baseline} & \textbf{0.0087} & \textbf{0.4814}
\\\textit{with uniform prior} & 0.0490 & 0.6648
\\ \textit{with cyclic LR} & 0.0491 & 0.7664
\\ \textit{with dropout} & 0.1171 & 1.0461
\\ \textit{with ResNet} & 0.1827 & 0.7238
\\
\bottomrule
\end{tabular}
\end{sc}
\end{small}
\end{center}
\end{table}

\subsection{Runtime}
Gaussianization Flows was by far the fastest model to train, requiring just 1.5 hours.
FFJORD required nearly two weeks to converge on our multimodal dataset. However, this extreme runtime may have in part been due to an issue with the reference implementation, which caused the ODE solver to run more slowly with each iteration.
WGAN with spectral normalization and 100 critic updates per generator update required around two days per run.
Hundreds of runs were necessary to find good hyperparameters.



\section{Discussion}
\label{discussion}
We developed synthetic datasets, reporting metrics, and a model-search methodology for evaluating both GANs and normalizing flows. Our results are surprising: GANs failed to learn key qualitative aspects of both unimodal and multimodal data. Quantitatively, normalizing flows outperformed the Wasserstein GAN in terms of the very metric that only latter targets: Wasserstein-1 distance. These negative results echo some concerns raised in~\citet{rosca2018distribution}.


There are caveats to our results. First, the lessons from low-dimensional data may not generalize to higher dimensional settings. However, for applications that require good recall, including many scientific applications, these results from low-dimensional data are enough to raise serious doubts about the performance of GANs in high dimensions, where there is no rigorous way to detect that an implicit model has poor recall. Good performance on our benchmarks is a necessary but not sufficient condition for a method to be a reliable at general-purpose density estimation.

Another caveat to our work is that establishing a general negative result requires exhaustively searching an infinite number of GANs, including GAN variants that have not yet been invented. We tested many but not all GAN variants. We invite others to test additional GAN variants using our benchmarking software, which is publicly available at \url{https://github.com/lliutianc/gan-flow}.
As it stands, however, scientists and other practitioners should be wary of relying on GANs to accurately estimate densities.





\bibliography{references}
\bibliographystyle{icml2020}

\newpage
\appendix

\end{document}